\pdfoutput=1

\documentclass[11pt]{article}

\usepackage[]{emnlp2021}

\usepackage{times}
\usepackage{latexsym}

\usepackage[T1]{fontenc}

\usepackage[utf8]{inputenc}

\usepackage{microtype}
\usepackage{hyperref}       
\usepackage{url}            
\usepackage{booktabs}       
\usepackage{amsfonts}       
\usepackage{nicefrac}       
\usepackage{algorithmicx}
\usepackage{algorithm}
\usepackage{algpseudocode}
\usepackage{amsmath} 
\usepackage{color}
\usepackage{graphicx}
\usepackage{subcaption}
\usepackage{multirow}
\usepackage{wrapfig}
\usepackage{latexsym}

\usepackage{times}
\usepackage{latexsym}
\usepackage{hyperref}
\usepackage{graphicx}
\usepackage{graphics}
\usepackage{multirow}
\usepackage{amsmath}
\usepackage{amsfonts}
\usepackage{amssymb}
\usepackage{tablefootnote}
\usepackage{color}
\usepackage{bbm}
\usepackage{graphicx}
\usepackage{xspace}
\usepackage{tikz}
\usepackage{pgfplots}
\usepackage[normalem]{ulem}
\usepackage{url}
\usepackage{subcaption}
\usepackage{fdsymbol}
\newcommand{\PET}{PET}
\newcommand{\La}{\mathcal{L}}
\newcommand{\ADAPET}{ADAPET}
\title{Prompt-based Text Entailment for Low-Resource Named Entity Recognition}

\author{Dongfang Li$^1$, Baotian Hu$^1$\footnotemark[1]\thanks{\hspace{2mm}Corresponding authors}\hspace{2mm},  Qingcai Chen$^{1,2}$\footnotemark[1]\hspace{1mm}\\
$^1$Harbin Institute of Technology (Shenzhen), Shenzhen, China \\
$^2$Peng Cheng Laboratory, Shenzhen, China\\
\texttt{crazyofapple@gmail.com, \{hubaotian, qingcai.chen\}@hit.edu.cn}}

\begin{document}
\maketitle
\begin{abstract}
Pre-trained Language Models (PLMs) have been applied in NLP tasks and achieve promising results. Nevertheless, the fine-tuning procedure needs labeled data of the target domain, making it difficult to learn in low-resource and non-trivial labeled scenarios. To address these challenges, we propose Prompt-based Text Entailment (\texttt{PTE}) for low-resource named entity recognition, which better leverages knowledge in the PLMs. We first reformulate named entity recognition as the text entailment task. The original sentence with entity type-specific prompts is fed into PLMs to get entailment scores for each candidate. The entity type with the top score is then selected as final label. Then, we inject tagging labels into prompts and treat words as basic units instead of n-gram spans to reduce time complexity in generating candidates by n-grams enumeration.  Experimental results demonstrate that the proposed method \texttt{PTE} achieves competitive performance on the CoNLL03 dataset, and better than fine-tuned counterparts on the MIT Movie and Few-NERD dataset in low-resource settings.

\end{abstract}

\section{Introduction}
Recently, Pre-trained Language Models (PLMs) have achieved promising improvement on several NLP tasks~\cite{bert,roberta,lan2019albert}. Nevertheless, fine-tuning language models still needs a moderate number of labeled data for downstream tasks. When difficulties result in limited labeled data available, the trained model shows large variance in downstream performance under full fine-tuning~\cite{DBLP:conf/iclr/MosbachAK21,huggface_prompt}. For example, labeling technical and professional terms can be time-consuming and labor-intensive in medical scenarios. Moreover, crowd-sourced annotation is also limited by the reality of existing samples (e.g., when online health assistants are applied to rare diseases).

To address learning challenges in these low-resource scenarios, researchers find that PLMs can learn well by prompt-based learning~\cite{schick2020exploiting,schick2020s,ADAPET}. Prompt-based learning models the probability of text directly; it does not need an extra fully-connected layer usually used by fine-tuning. The main idea is to reformulate NLP tasks as cloze-style question answering for better using the knowledge in PLMs. The model predicts the word probability of masked positions and then derives the final output via mapping relations between words and labels. Previous works have shown the ability of prompt-based learning under low-resource settings~\cite{schick2020exploiting,schick2020s,schick-etal-2020-automatically,Lester}. For example,
some prompt-based works have explored in classification and generation tasks where it is relatively easy to reformulate into cloze-style tasks (cf. Section~\ref{sec:related}). Nevertheless, the application to Named Entity Recognition (NER) still poses challenges for current methods. Unlike text classification and text generation, NER is the task of identifying named entities (e.g., \textit{person name}, \textit{location}) in a given sentence, and each unit of the input needs to be predicted. If we directly use Masked Language Modeling (MLM) head to predict each unit label, the lexical and semantic coherence are ignored as there exists latent relationships between the tokens~\cite{lample2016neural,elmo,TENER}.

In this work, we propose \textbf{Prompt-based Text Entailment} (\texttt{PTE}) for low-resource NER. Firstly, we reformulate NER as a \textit{text entailment} task.
Textual Entailment (TE) is the task of studying the relation of two sentences, Premise (P) and Hypothesis (H): whether H is true given P~\cite{snli}. Specifically, we treat the original sentence as premise and entity type-specific prompt as a hypothesis. Given an entity type, the P and H are fed into PLMs to get entailment scores for each candidate. Then, the entailment score is the probability of a specific token at the mask position of the prompt. After that, the entity type with the top entailment score is selected as the final label.  
During inference, we enumerate all possible text spans or words in the input sentence as named entity candidates~\cite{templatener}. The reformulation provides a unified entailment framework for NER tasks where annotations are insufficient, as the model shares the same inference pattern across different domains. As such, we can also leverage generic text entailment datasets such SNLI~\cite{snli} and MNLI~\cite{mnli} to pre-train models, which transfer knowledge from the general domain and get better performance in new domains. Our method can be a step forward towards the development of a solution for the low-resource NER because any new domain does not typically have extensive annotated data in the real world, whereas it is feasible to obtain a couple of examples (e.g., online assistant). Moreover, considering the existence of noisy annotations, \textit{TE only needs to specify the labels of certain entities for training rather than the complete annotations of the entire sequence}.
Experimental results demonstrate that the proposed method \texttt{PTE} achieves competitive F1 score on the CoNLL03 dataset~\cite{conll03}, and better than fine-tuned counterparts by a large margin on the MIT Movie~\cite{mit-dataset} and Few-NERD datasets~\cite{fewnerd} in low-resource settings.

\section{Method}

\subsection{Low-Resource Named Entity Recognition}
\label{sec:fewshot}
Given a sentence $\mathbf{X} = (x_{1}, x_{2}, \dots x_{N})$ which contains $N$ words, the task is to produce $\mathbf{Y} = (y_{1}, y_{2}, \dots y_{N})$ which is the sequence of entity tags. The tag $y_{i}\in\mathcal{Y}$ (e.g., B-LOC, I-PER, O) denotes the type of entity for each word $x_{i}$, where $\mathcal{Y}$ is a pre-defined set of tags.
We are given a low-resource NER dataset $\mathcal{D}_{\text{train}}$, where the labeled examples to each NER type (e.g., $<50$) are substantially less than that in the rich-resource NER dataset. Our goal is to train an accurate NER model under this low-resource setting.

Previous methods usually treat NER as a sequence labeling task in which a neural encoder such as LSTM and BERT is used for representing the input sequence, and a softmax or a CRF layer is equipped as the output layer to get the tag sequence.
Formally, as the standard fine-tuning, NER model $\mathcal{M}$ parameterized by $\theta$ is trained to minimize the cross-entropy loss over token representations $\mathbf{H} = [h_{1}, h_{2}, \dots h_{N}]$ that are generated from the neural encoder as follows:
\begin{equation}
\mathcal{L}=-\sum_{i=1}^{N} \log f_{x_{i}, y_{i}}(\mathbf{h} ; \theta),
\end{equation}
where $f$ is the model's predicted conditional probability for golden label.

\subsection{Prompt-based Text Entailment}
Towards the low-resource NER task, a common way is to pre-train the neural encoder and output layer parameters with the rich-resource NER dataset. Another feasible way is to focus on the matching function learned by prototype-based network~\cite{proto1} or nearest neighbor classification~\cite{proto2}. After that, a well-trained matching function can work well in the target tasks.
However, since the entity category is different, the parameter for the low-resource domain cannot be transferred directly from the source domain. Moreover, the metric-based meta-learning methods assume that training and test tasks are in the same distribution but this assumption may not always be satisfied in practice~\cite{yin2020meta}. 

In this work, we reformulate named entity recognition as the text entailment task. As the NER task is not a standard entailment problem, we convert NER examples into labeled entailment instances. The input includes the original sentence as premise and entity type-specific prompt as a hypothesis (i.e., template). The output is produced by an entailment classifier, predicting a label for each instance.
The entailment score is the probability of a specific token at the mask position of the prompt. Then, the entity type with the top entailment score is selected as the final label. For example, given a sentence ``\textit{Seoul is the capital of South Korea.}'' and a candidate ``\textit{Seoul}'', we define ``\textit{Seoul is an <entity\_type> entity. [MASK]}'' as prompt for each entity type. Suppose the entailment score of token ``\textit{yes}'' at [MASK] for <location> type is the highest 
of all entity types, we finally choose ``\textit{location}'' as the predicted label. 
For training, we sample three types of negative examples (see Appendix \ref{sec:templ}): false positive (i.e., replace the correct label with others), null label (i.e., replace the correct label with null), and non-entity replacement (i.e., replace golden entity with non-entity span). For example,  ``\textit{Seoul is not a named entity. [MASK]}'' is one prompt of ``false positive'' example (i.e., the [MASK] label is \textit{no}, and it exists entities). During training and inference, we can enumerate all possible text spans in the input sentence as named entity candidates~\cite{templatener}. 
To further reduce time complexity in generating candidates by n-grams enumeration, we inject tagging labels (e.g., I-location means the tag is inside a entity) into prompts and treat words as basic units instead of text spans during training and inference. In other words, we consider prompts ``\textit{<candidate\_entity\_word> is the part of a <entity\_type> entity. [MASK]}'' (e.g., ``\textit{Seoul is the part of a location entity. [MASK]''}). As PTE treats words as basic units for decoding, it optimizes time complexity at inference to O(L), which is in line with previous NER methods. It optimizes quadratic costs at inference to linear. We also apply the Viterbi algorithm at inference, where transitions are computed on the training set \cite{DBLP:conf/acl/HouCLZLLL20}. The computational complexity of n-grams enumeration is O($L^2$), increasing quadratically with sequence length L.  Overall, our method provides a unified entailment framework as the model shares the same inference pattern across different domains.

\subsection{Pattern Exploiting Training Framework} \label{sec:pet}
The basic framework of \texttt{PTE} is from {\ADAPET} \cite{ADAPET} which is a variant of  {\PET} \cite{schick2020exploiting, schick2020s}. Compared with {\PET}, {\ADAPET} uses more supervision by decoupling the losses for the label tokens and a label-conditioned MLM objective over the total original input~\cite{ADAPET}.  We introduce it by describing how to convert one example into a cloze-style question. The query-form in {\ADAPET} is defined by a Pattern-Verbalizer Pair (PVP). Each PVP consists of one pattern which describes how to convert the inputs into a cloze-style question with masked out tokens, and one verbalizer which describes the way to convert the classes into the output space of tokens. The PVP can be manually generated \cite{auxiliary-absa,lama} or obtained by using an automatic search algorithm \cite{schick-etal-2020-automatically, gao2020making}. 
After that, {\ADAPET} obtains logits from the model $G_m(x)$. Given the space of output tokens $\mathcal{Y}$, {\ADAPET} computes a softmax over $y \in \mathcal{Y}$, using the logits from $G_m(x)$. The final loss is shown as follows: 

{\small
\begin{align}
    q(y|x) &= \frac{\exp([\![G_m(x)]\!]_{y})}{\sum\limits_{y' \in \mathcal{Y}} \exp([\![G_m(x)]\!]_{y'})}, \\
    \La &= \texttt{Cross\_entropy} (q(y^*|x), y^*). \ 
    \label{eq:pet}
\end{align}
}%

\subsection{Cross Task and Domain Transfer}
\label{sec:template}
To address the challenge when few labeled examples are available, we further train the sentence encoder on the TE datasets (e.g., MNLI) and apply it to the NER task. Then, our method can perform more knowledge transfer between the rich-resource NER dataset and the low-resource NER dataset. Since there is no domain-related fully connected layer for fine-tuning, all parameters can be transferred in different domains even if the entity category does not match. Specially, we apply the text entailment method to the low-resource domain after firstly pre-training the NER model in the rich-resource domain. This process is simple but can effectively transfer label knowledge. As the output of our method is model-agnostic words (not tag index), the tag vocabulary with rich-resource and low-resource is a shared pre-trained language model vocabulary set. It allows our method to use the correlation of tags to enhance the effect of cross-domain transfer learning.

\section{Experiments}
We compare our methods with several baselines on both rich-resource settings and low-resource settings. We use the CoNLL2003 \cite{conll03} as the rich-resource dataset, and MIT Movie \cite{mit-dataset}, Few-NERD \cite{fewnerd} as the cross-domain low-resource datasets. And we conduct experiments on the CoNLL03 dataset in both full and low-resource settings. The dataset statistics and experimental settings are included in Appendix~\ref{sec:dataset} and~\ref{sec:exp_settings}. The standard precision, recall, and F1 score are used for model evaluation.

\subsection{Rich-Resource NER Results}

We first use the whole training set of the CoNLL03 to train the model and evaluate its performance on the test set. 
Table~\ref{table:mp_conll} shows the performance of the comparison method and our model on the test set. We can find that although the potential applications of \texttt{PTE} is low-resource named entity recognition, it can also achieve competitive performance in rich-resource domain data sets. Compared with BERT fine-tuning reported in the previous work, the \texttt{PTE} model using discrete manual design reduces the F1 by 0.32, while the \texttt{PTE} model using the soft prompt method design mode~\cite{liu2021ptuning,ptuning} increases the F1 by 0.5. It shows that our method effectively recognizes named entities, and soft prompts can improve performance compared with manually designed prompts. More experimental results about TE patterns (\S\ref{sec:pet}) are in the Appendix~\ref{sec:pattern}.

\subsection{Cross Entity Type NER Results} 
\label{cross_type}

Following~\citet{templatener}, we sample the number of examples corresponding to different types of entities on the CoNLL03 data training set as new training set while keep test set unchanged. Among them, ``PER'' and ``ORG'' are rich-resource entity types, and ``LOC'' and ``MISC'' are low-resource entity types. The experimental results are shown in Table~\ref{table:conll_few}. The results show that our method achieves better results than baselines on the low-resource entity types, thus improving overall performance. On the other hand, our method is better than fine-tuning in both cases.

\begin{table}[t!]
\centering
\small
\resizebox{\linewidth}{!}{
\begin{tabular}{l|c|c|c}
     \hline
    {\bf Method} & {\bf Precision} & {\bf Recall} & {\bf F1} \\
    \hline
    \citet{label-agnostic} & - & - & 89.94 \\
    \citet{yang-etal-2018-design} & - & - & 90.77 \\
    \citet{lstm-cnn-crf} & - & - & 91.21 \\
     BERT~\cite{templatener} & 91.93 & 91.54 & 91.73 \\
    \citet{DBLP:conf/emnlp/YamadaASTM20} & - & - & \bf{94.30} 
     \\
     \hline
     Template BART~\cite{templatener} & 90.51 & 93.34 & 91.90 \\
     
     \texttt{PTE} (discrete) & 91.27 & 91.56 & 91.41 \\
     \texttt{PTE} (soft) & 92.01 & 92.45 & \bf{92.23} \\
     \hline
\end{tabular}
}
\caption{Model performance on the CoNLL03 test set.\label{table:mp_conll}}
\end{table}

\begin{table}[t]
\centering
\small
\resizebox{\linewidth}{!}{

\begin{tabular}{l|c|c|c|c|c}
     \hline
   \textbf{Method} & {\bf PER} & {\bf ORG} & {\bf LOC} & {\bf MISC} & {\bf Overall} \\
    \hline
    BERT & 75.71 & 77.59 & 60.72 & 60.39 & 69.62 \\
    Template BART & 84.49 & 72.61 & 71.98 & 73.37 & 75.59 \\
    \texttt{PTE} (BERT) & 85.34 & 72.89 & 73.01 & 74.32 & \bf{76.40} \\
     \hline
\end{tabular}
}
\caption{Cross entity type results on the CoNLL03. LOC and MISC are low-resource entity types, where PER and ORG are rich-resource entity types. \label{table:conll_few}}
\end{table}
\subsection{Domain Transfer for Low-Resource NER}

\begin{table}[t!]
\centering
\small

\vspace{-0.2cm}
\resizebox{\linewidth}{!}{
\begin{tabular}{l|c|c|c|c|c|c}\hline
\multicolumn{7}{c}{\textit{MIT Movie} (12)}\\
\hline

     \textbf{Method} & {\bf K=10} & {\bf K=20} & {\bf K=50} & {\bf K=100} & {\bf K=200} & {\bf K=500} \\
\hline
    \citet{label-agnostic} &\ 3.1 &\ 4.5 &\ 4.1 &\ 5.3 &\ 5.4 &\ 8.6 \\
    \citet{example-ner} & 40.1 & 39.5 & 40.2 & 40.0 & 40.0 & 39.5 \\
    Sequence Labeling BERT & 28.3 & 45.2 & 50.0 & 52.4 & 60.7 & 76.8 \\
    \citet{DBLP:conf/emnlp/YamadaASTM20} & 35.6 & 49.2 & 61.8 & 72.4 & 78.7 & 82.8 \\
    Template BART~\cite{templatener} & 42.4 & 54.2 & 59.6 & 65.3 & 69.6 & 80.3 \\ \hline
     \texttt{PTE} (discrete) & 46.9$\dagger$ & 59.2$\dagger$ & 66.9$\dagger$ & 74.9$\dagger$ & 79.9$\dagger$ & 83.6\\
     \texttt{PTE} (soft) & \textbf{47.8}$\dagger$ & \textbf{60.8}$\dagger$ & \textbf{68.1}$\dagger$ & \textbf{76.5}$\dagger$ & \textbf{83.6}$\dagger$ & \textbf{86.4}$\dagger$ \\
    \hline \hline
    \multicolumn{7}{c}{\textit{Few-NERD} (8)}\\\hline 

     \textbf{Method} & {\bf K=10} & {\bf K=20} & {\bf K=50} & {\bf K=100} & {\bf K=200} & {\bf K=500} \\
\hline
     \citet{label-agnostic} &\ 5.2 &\ 4.1 &\ 4.7 &\ 7.8 &\ 12.3 &\ 10.1 \\
     \citet{example-ner} & 35.4 & 48.3 & 51.2 & 51.8 & 53.6 & 55.7 \\
     Sequence Labeling BERT & 50.6 & 59.3 & 61.3 & 61.4 & 62.5 & 66.4 \\ 
     \citet{DBLP:conf/emnlp/YamadaASTM20} & 51.7 & 60.1 & 62.3 & 61.0 & 62.5 & 66.8 \\ 
     \hline
     \texttt{PTE} (discrete) & 51.8 & 59.7 & 60.5 & 61.3 & 61.8 & 63.4\\
     \texttt{PTE} (soft) & \textbf{54.2} & \textbf{61.4} & \textbf{62.3} & \textbf{62.5} & \textbf{63.6} & \textbf{67.4} \\

\hline
\end{tabular}
}
\caption{\label{tab:fewshot}F1 comparison of two low-resource NER datasets. We set 6 sample size $K$ for different low-resource settings. $\dagger$ means a significant difference compared to Template BART ($p < .05$).}
\vspace{-0.2cm}
\end{table}

We do not use $N$-way $K$-shot setting~\cite{proto2,fewnerd} which samples $N$ categories and $K$ examples for training in each episode because a sentence in the NER task may contain multiple entities from different types. Thus, we randomly sample training data from the MIT Movie and Few-NERD datasets to simulate low-resource scenarios and use CoNLL03 as the rich-resource dataset. As such, we have only $K$ examples for each type of training. We choose $K\in\{10,20,50,100,200,500\}$ for experiments to evaluate the ability of the model on training data of different sizes. The experimental results are in Table~\ref{tab:fewshot}. The results show that when the $K$ value is relatively small, our \texttt{PTE} method can be better than the fine-tuning method, and this trend decreases with the increase of $K$. In addition, the soft mode is also better than the discrete mode in the case of a small number of samples. Overall, our method achieves the best results on both data sets in the low-resource scenario.

\begin{figure}[t!]
\centering
\includegraphics[width=1.0\linewidth]{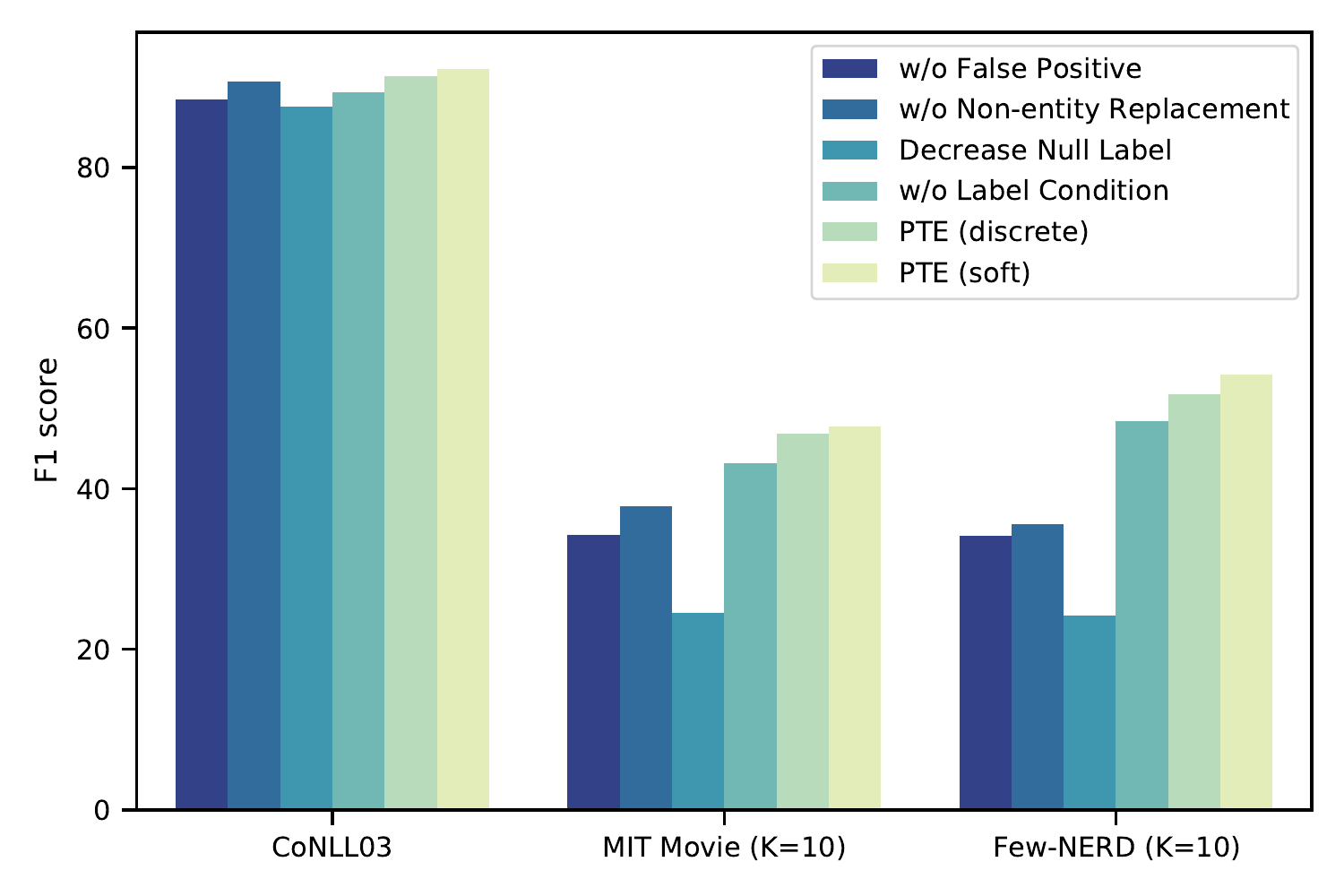}
  \caption{F1 scores with different experimental settings and model variants.}
  \vspace{-4mm}
   \label{figablition}
\end{figure}
\subsection{Ablation Study}

We conduct ablation experiments and the results are shown in Figure~\ref{figablition}. The results show that (1) the selection of negative examples has a great impact on the performance of the model, especially the negative examples of the null label type. However, in rich-resource scenario, the gap between full setting and decreased setting is not as much as the low-resource scenario; (2) the low-resource scenario is a challenge to the model, and the results of some variants are not inconsistent where prompt-based learning may not be as good as fine-tuning; (3) label conditioning and soft mode have a consistent effect on the model. These findings highlight that it still has room left to use prompt for effectively transferring knowledge in the case of low-resource scenario.
\section{Related Work}
\label{sec:related}
Previous works have shown the ability of prompt-based learning under low-resource settings\cite{schick2020exploiting,schick2020s,schick-etal-2020-automatically,Lester}. \citet{schick2020exploiting} address low-resource text classification by manually designing templates as prompt-based learning in a iterative training manner. \citet{gao2020making} improve low-resource performance with well-designed templates with demonstrations. \citet{ptuning} apply continuous prompts for low-resource learning. Recently, some works~\cite{fewnerd,tong-etal-2021-learning,tempfree,LightNER} also focus on low-resource NER.  In contrast, we propose to use prompt-tuning to treat NER as the TE task. Unlike traditional NER methods, we use prompt-based learning without an additional linear layer for fine-tuning. By defining different prompts, the model is able to perform well in low-resource settings, which adapts to new domains with few labeled data. In contrast to recent work which also adopts prompt-based fine-tuning for NER ~\cite{DBLP:journals/corr/abs-2109-13532}, we show that the effectiveness of the text entailment reformulation for named entity recognition using PLMs.
\section{Conclusion}
In this paper, we apply prompt-based learning to low-resource named entity recognition. For token classification of NER, we reformulate it into a text entailment task. Our method transfers knowledge in different NLP tasks and domains, and performs better in low-resource scenarios. Future work includes how to apply \texttt{PTE} to other NLP tasks.

\section*{Acknowledgements}
We thank the anonymous reviewers for their insightful comments and suggestions. This work is jointly supported by grants: Natural Science Foundation of China (No. 62006061,61872113,U1813215), Stable Support Program for Higher Education Institutions of Shenzhen (No. GXWD20201230155427003-20200824155011001) and Strategic Emerging Industry Development Special Funds of Shenzhen(No. JCYJ20200109113441941 and No. XMHT20190108009).
\bibliography{anthology,custom}
\bibliographystyle{acl_natbib}
\clearpage


\begin{table}[t!]
\centering
    \resizebox{\linewidth}{!}{
\begin{tabular}{llccccc}
\toprule
\textbf{Datasets} & \textbf{Domain} & \textbf{\# Type} &  \textbf{\# Tokens} & \textbf{\# Train} & \textbf{\# Dev} & \textbf{\# Test} \\
\midrule
CoNLL03 & Reuters news stories & 4 & 21.0k & 14041 & 3250 & 3453 \\
MIT Movie & Movie reviews & 12 & 6.0k & 7820 & 1955 & 2443\\
Few-NERD & Wikipedia & 8 & 4601.2k & 131767 & 18824 & 37648 \\
\bottomrule

\end{tabular}}
\caption{Statistics of our datasets. We count the number of sentences in the training/development/test set, the number of tokens and the number of tags in datasets.}
\label{tab:dataset}
\end{table}

\begin{table}[t!]
\small
    \centering
    \resizebox{\linewidth}{!}{
    \begin{tabular}{l|l}
    \hline
        \textbf{Type} & \textbf{Templates} \\ 
    \hline
     Positive (Y) & <candidate> is the part of a <entity\_type> entity. \\
     \hline
     False positive (N) & <candidate> is the part of a <another\_entity\_type> entity. \\
     \hline
     Non-entity (N) &  <others> is the part of a <entity\_type> entity. \\
     \hline
      \multirow{2}{*}{Null label (Y/N)} & <others> is not a name entity. \\
      & <candidate> is not a name entity.\\
     \hline
 
    \end{tabular}
    }
    \caption{The discrete manually-crafted templates.}
    \label{tab:template}
\end{table}

\begin{table}[t!]
\small
    \centering
    \resizebox{\linewidth}{!}{
    \begin{tabular}{l|l}
    \hline
      \textbf{Number}  & \textbf{Patterns} \\ 
    \hline
     Pattern\#1 & [HYPOTHESIS] ? </s></s> [MASK], [PREMISE] </s> \\
     \hline
      Pattern\#2 & `` [HYPOTHESIS] '' ? </s></s> [MASK], `` [PREMISE] '' </s> \\
     \hline
      Pattern\#3 & [HYPOTHESIS] ? </s></s> [MASK]. [PREMISE] </s> \\ 
     \hline
       Pattern\#4 & `` [HYPOTHESIS] '' ? </s></s> [MASK]. `` [PREMISE] '' </s> \\
     \hline
 
    \end{tabular}
    }
    \caption{We list the patterns used by our method where <s> and </s> are start token and separated token.}
    \label{tab:pattern}
\end{table}
\begin{figure}[t!]
\centering
\includegraphics[width=0.8\linewidth]{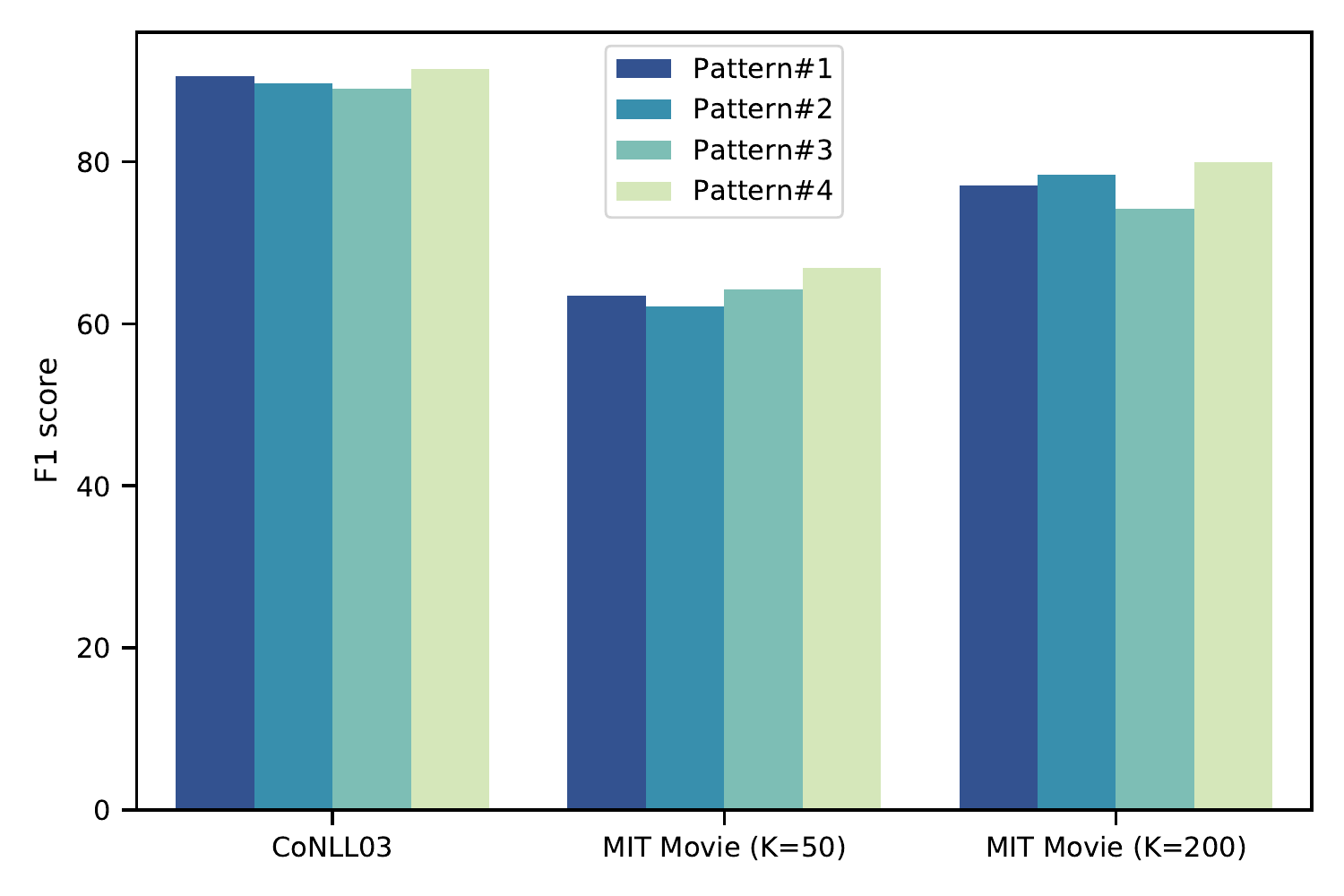}
  \caption{The performance of different modes after up to 7000 training batches. The patterns we use are from RTE task of \citet{ADAPET}. }
  \vspace{-4mm}
   \label{figpatt}

\end{figure}
\appendix
\label{sec:supplemental}

\section{Templates}
\label{sec:templ}

We use the naive random sampling method and the positive-negative ratio is 1:1.5 in the low-resource scenario after sampling. 
As shown in Table~\ref{tab:template}, we list our templates for each example type used by \texttt{PTE} (discrete). Our soft prompt is to add different special tokens before the [MASK] to form a template and tune the embeddings of these tokens directly following~\citet{typing} used by \texttt{PTE} (soft). We leave it for future work to examine whether the NER performance further improves with a more well-designed soft prompt.

\section{Dataset Statistics}
\label{sec:dataset}
We use the following datasets where data statistics are displayed in Table~\ref{tab:dataset}: (1) The CoNLL03 dataset~\cite{conll03} is from the English Reuters News and consists of 4 entity types. We use the previous split in~\citet{templatener} for our experiments. The entity types are \textit{person}, \textit{location}, \textit{organization}, and \textit{miscellaneous entities}. The sampled training dataset in \S \ref{cross_type} includes 1500 organization entities, 1500 person entities, 150 location entities and 150 miscellaneous entities  (2) The MIT Movie dataset~\cite{mit-dataset} is from queries related to movie information. The entity types are \textit{actor}, \textit{character}, \textit{director}, \textit{genre}, \textit{plot}, \textit{year}, \textit{soundtrack}, \textit{opinion}, \textit{award}, \textit{origin}, \textit{quote}, and \textit{relationship}. (3) The Few-NERD dataset~\cite{fewnerd} is a low-resource NER dataset with a hierarchy of 8 coarse-grained and 66 fine-grained entity types. We use the coarse-grained entity in our experiments. The entity types are \textit{location}, \textit{event}, \textit{building}, \textit{art}, \textit{product}, \textit{person}, \textit{organization}, and \textit{miscellaneous entities}.

\section{Experimental Settings}
\label{sec:exp_settings}
We use the pre-trained models and codes provided by ADAPET and follow their default hyperparameter settings unless noted otherwise. The pre-trained language model of our method is BERT that is pre-trained in the MNLI datasets. We use AdamW optimizer and grid search batch size of \{$8$,$16$,$32$\} for model training.  We use grid search for learning rate from $[1\text{e-}5, 2\text{e-}5, 3\text{e-}5, 4\text{e-}5, 5\text{e-}5]$. And we grid search the optimal weight decay weight from $[0.1, 0.01, 0.005, 0.001]$. The maximum sequence length, the dropout rate, the gradient accumulation steps, the maximum training steps and the warm-up ratio are set to $256$, $0.1$, $16$, $7000$, $0.06$ respectively. Early stopping is also applied based on model performance on the development set. Our models are trained with NVIDIA Tesla V100s. The verbalizer words are [``yes'', ``no''] and [``true'', ``false'']. The $\tau$ of transition probability in decoding is selected by searching with $0.05$ step from $0$ to $1$. For sequence labeling BERT fine-tuning, we train BERT with a softmax classifier following \citet{bert}, updating parameters using Adam with an initial learning rate of $1\text{e-}5$, and a batch size of $32$. 

\section{Pattern Engineering}
\label{sec:pattern}

After designing templates of entity-specific hypothesis, we follow \citet{ADAPET} to define the TE patterns in Table~\ref{tab:pattern} and report results across all patterns for all datasets in Figure~\ref{figpatt}. We find that the subtle difference of the prompts impacts performance, while Pattern\#4 outperforms others across datasets and settings.

\end{document}